\documentclass{article}
\usepackage{iclr2018_conference,times}
\iclrfinalcopy 

\usepackage[utf8]{inputenc} 
\usepackage[T1]{fontenc}    
\usepackage{hyperref}       
\usepackage{url}            
\usepackage{booktabs}       
\usepackage{amsfonts}       
\usepackage{nicefrac}       
\usepackage{microtype}      
\usepackage{amsmath}
\usepackage{algorithm}
\usepackage{algpseudocode}
\usepackage{graphicx}
\usepackage{todonotes}
\usepackage{xargs}
\usepackage{booktabs}

\providecommand{\scan}{\text{SCAN}}
\providecommand{\reduce}{\text{REDUCE}}

\algblock{ParFor}{EndParFor}
\algnewcommand\algorithmicparfor{\textbf{parfor}}
\algnewcommand\algorithmicpardo{\textbf{do}}
\algnewcommand\algorithmicendparfor{\textbf{end\ parfor}}
\algrenewtext{ParFor}[1]{\algorithmicparfor\ #1\ \algorithmicpardo}
\algrenewtext{EndParFor}{\algorithmicendparfor}

\newcommandx{\Chris}[2][1=]{\todo[linecolor=red,backgroundcolor=red!25,bordercolor=red,#1]{#2}}
\newcommandx{\Eric}[2][1=]{\todo[linecolor=blue,backgroundcolor=blue!25,bordercolor=blue,#1]{#2}}
\newcommandx{\thiswillnotshow}[2][1=]{\todo[disable,#1]{#2}}

\title{Parallelizing Linear Recurrent Neural Nets Over Sequence Length}

\author{Eric Martin \\
\texttt{eric@ericmart.in} \\
\And
Chris Cundy \\
Department of Computer Science\\
University of California, Berkeley\\
Berkeley, CA 94720, USA\thanks{Currently at the Future of Humanity Institute, University of Oxford, Oxford, UK} \\
\texttt{c.cundy@berkeley.edu} \\
}

\begin{document}
\maketitle

\begin{abstract}
Recurrent neural networks (RNNs) are widely used to model sequential data but
their non-linear dependencies between sequence elements prevent parallelizing
training over sequence length. We show the training of RNNs with only linear
sequential dependencies can be parallelized over the sequence length using the
parallel scan algorithm, leading to rapid training on long sequences even with
small minibatch size. We develop a parallel linear recurrence CUDA kernel and
show that it can be applied to immediately speed up training and inference of
several state of the art RNN architectures by up to 9x.  We abstract recent work
on linear RNNs into a new framework of linear surrogate RNNs and develop a
linear surrogate model for the long short-term memory unit, the GILR-LSTM, that
utilizes parallel linear recurrence.  We extend sequence learning to new
extremely long sequence regimes that were previously out of reach by
successfully training a GILR-LSTM on a synthetic sequence classification task
with a one million timestep dependency.
\end{abstract}

\section{Introduction}
Recurrent neural networks (RNNs) are widely used for sequence modelling tasks in
domains such as natural language processing \citep{sutskever2014sequence},
speech recognition \citep{amodei2015deep}, and reinforcement learning
\citep{hausknecht2015deep}. Most RNNs, including popular variants such as long
short-term memories (LSTMs), introduced by \citet{hochreiter1997long}, and gated
recurrent units (GRUs), introduced by \citet{cho2014learning}, contain a non-linear dependency
between sequential inputs. These non-linear dependencies create a very flexible
class of models but limit the feasibility of training RNNs on long sequences as
each sequence element must be processed sequentially.  Modelling sequences of
thousands to millions of elements is important to domains such as robotics,
remote sensing, control systems, speech recognition, medicine, and finance.

The RNN serial evaluation inefficiency problem is usually mitigated by
parallelizing the forward and backward pass over a minibatch of inputs. Without
minibatches, RNN evaluation is a sequence of matrix-vector
multiplications. Minibatches transform RNN computation into a sequence of more
efficient matrix-matrix multiplications, but this speed-up brings several
disadvantages.  RNN model size is often limited by GPU memory size, and running
a forward and backward pass on a minibatch requires memory linear in the
minibatch size.  Grouping data into minibatches increases the latency of each
pass and reduces the rate of optimization steps. Finally, training with larger
minibatches damages generalization ability \citep{keskar2017large}. Given these
effects, it is desirable to obtain high training throughput with small minibatches.
Persistent RNNs \citep{diamos2016persistent} use a novel
implementation that can achieve high GPU utilization with very small minibatch
sizes when the recurrent state is larger than 500 elements, but even persistent
RNNs become limited by the serial evaluation inefficiency at smaller hidden
sizes.

Numerous prior works have shown strong performance from neural sequential models
with only linear dependence on earlier sequence
elements. \citet{balduzzi2016strongly} investigated RNNs with only elementwise
linear recurrence relations $h_t = \alpha_t \odot h_{t-1} + (1-\alpha_t) \odot
x_t$ and developed linear variants of LSTM and GRU that perform similarly to
standard non-linear RNNs on text generation tasks. \citet{bradbury2017quasi},
\citet{kalchbrenner2016neural}, \citet{gehring2017convolutional}, and
\citet{van2016wavenet} have successfully applied networks of convolutions over
sequences for tasks such as machine translation, language modelling, and audio
generation.  These works have observed up to an order of magnitude increase in
training throughput compared to RNN alternatives. Convolutional sequence models
typically rely on either an attention mechanism or a (possibly linear) recurrent
layer to integrate information at scales larger than the filter
width. Introduction of a recurrent layer prevents full parallelization over the
sequence length while attention mechanisms are expensive to apply on long
sequences in online inference use cases.

A linear recurrence is a specific instance of a general form of computation known as a scan.
Scans and reductions are computations involving repeated application of a binary
operator $\oplus$ over an array of data. Computing the sum or maximum
of an array is an example of a reduction, while a cumulative sum is a common
example of a scan operation. Throughout this work, the scan of $\oplus$ with
initial value $b$ is defined as
\begin{align*}
\scan(\oplus, [a_1, a_2, ..., a_n], b) = [(a_1 \oplus b), (a_2 \oplus a_1 \oplus b), ..., (a_n \oplus a_{n-1} ... \oplus a_1 \oplus b)].
\end{align*}
The reduction of $\oplus$ over array $A$ and initial value $b$ is denoted
$\reduce(\oplus, A, b)$ and is the final element of $\scan(\oplus, A, b)$.
Despite their dependent computation graph, algorithms exist to parallelize scans
and reductions when $\oplus$ is associative \citep{ladner1980parallel}.

\citet{blelloch1990prefix} shows that first order recurrences of the form
$h_t = (\Lambda_t \otimes h_{t-1}) \oplus x_t$ can be parallelized with
the parallel scan algorithm if three conditions are met:

\begin{enumerate}
\item $\oplus$ is associative: $(a \oplus b) \oplus c = a \oplus (b \oplus c)$
\item $\otimes$ is semiassociative: there exists a binary associative operator
$\odot$ such that $a \otimes (b \otimes c) = (a \odot b) \otimes c$
\item $\otimes$ distributes over $\oplus$: $a\otimes(b\oplus c) = (a\otimes b) \oplus (a \otimes c)$
\end{enumerate}

Considering the familiar operations in linear algebra, we see that the
associative operation of vector addition
($x \oplus y = x + y$), the
semiassociative operation of matrix-vector multiplication
($A \otimes x = Ax$) and the associative operation of
matrix-matrix multiplication ($A \odot B=AB$) satisfy Blelloch's three
conditions, allowing $h_t = \Lambda_t h_{t-1} + x_t$ to be evaluated in parallel
over time steps \(t\) for vectors $x_t$ and square matrices $\Lambda_t$.

We investigate this idea further and deliver the following contributions:
\begin{itemize}
\item{We classify RNNs which satisfy the conditions above, and show that many
    RNNs used in practice such as the Quasi-RNNs (QRNNs) introduced by
    \citet{bradbury2017quasi} are contained in this class.}
\item{We provide an implementation of the parallel linear recurrence algorithm
    as a CUDA kernel, and show that it speeds up training of QRNN and
    \citet{lei2017}'s Simple Recurrent Unit (SRU) architectures by factors of up
    to 9x.}
\item{We describe how several recent linear RNNs can be described as
    linear surrogates for non-linear architectures. We introduce a linear
    surrogate for the LSTM and show that we are able to train it
    with a speedup of 5-10x compared to the CuDNN LSTM when we use the
    parallel linear recurrence algorithm.}
\end{itemize}

\section{Parallel linear recurrence}
As the method is
essential to this work, Algorithm 1 presents the parallel linear recurrence
algorithm for the interested reader.
\begin{algorithm}
  \label{alg:plr}
\caption{Parallel linear recurrence on $p$ processors}
\begin{algorithmic}[1]
  \State Let $y = [(\Lambda_1, x_1), (\Lambda_2, x_2), ..., (\Lambda_T, x_T)]$
  \State Let binary operator $\bullet$ act as $(\Lambda, x) \bullet h = \Lambda h + x$
  \State Let $S_0=1, S_i < E_i, E_i + 1 = S_{i+1}, E_{p-1}=T$ for $i$ in $0,p-1$

  \\
  \ParFor{$i \gets 0,p-1$}
    \State $P_i = \reduce(\odot, \Lambda_{S_i:E_i}, I)$
    \State $R_i = \reduce(\bullet, y_{S_i:E_i}, 0)$
  \EndParFor

  \\
  \State Let $z = [(P_0, R_0), (P_1, R_1), ..., (P_p, R_p)]$.
  \State $C = \scan(\bullet, z, h_0)$   \Comment{compute $C_i = P_i C_{i-1} + R_i$ with $C_{-1}=h_0$}

  \\
  \ParFor{$i \gets 0,p-1$}
    \State $h_{S_i:E_i} = \scan(\bullet, y_{S_i:E_i}, C_{i-1})$
  \EndParFor

  \State \Return $h$
\end{algorithmic}
\end{algorithm}

\subsection{Theoretical performance}
The cost of a serial scan over a sequence of length $T$ is
$C_\text{sscan} \in \mathcal{O}((C_\otimes + C_\oplus)T)$, compared to the parallel scan cost
$C_\text{pscan} \in \mathcal{O}(2(C_\odot + C_\otimes + C_\oplus)(T/p + \lg p))$ on $p$ processors \citep{blelloch1990prefix}.
If $h_t$ is a vector of dimension $n$ then
$C_\odot \in  \mathcal{O}(n^3), C_\otimes \in \mathcal{O}(n^2), C_\oplus \in \mathcal{O}(n)$ giving
$C_\text{pscan} \in \mathcal{O}(2(n^3 + n^2 + n)(T/p + \lg p))$ and
$C_\text{sscan} \in \mathcal{O}((n^2 + n)T)$. The $\mathcal{O}(n^3)$ cost of the matrix
multiplication in the parallel algorithm can counter-act any parallel speedups for
sufficiently large hidden states and lead to a slower algorithm overall.

To avoid this problem, we will only consider diagonal matrices $\Lambda_t$, in
which case both matrix-matrix and matrix-vector multiplication have cost
proportional to $n$ and $C_\text{pscan}\in \mathcal{O}(6n(T/p + \lg p))$ and
$C_\text{sscan} \in \mathcal{O}(2nT)$. This gives a parallel speedup factor of
$pT/3(T+\lg p)$. Assuming $p \ll T$, then $C_\text{pscan} \le
C_\text{sscan}$ when $p \ge 3$.

As we are only considering diagonal matrices,
we write the linear recurrence as $h_t = \lambda_t \odot h_{t-1} + x_t$
where $\odot$ indicates elementwise multiplication.

Limiting $\Lambda_t$ to be diagonal may seem like a severe constraint but there are
several reasons to do so beyond the favorable parallelization performance. Relatively few neural
network models use separate recurrent matrices for each sequence element and using these
separate matrices would require potentially prohibitive $n^2T$ memory.
Applying
the same matrix $\Lambda$ to each sequence element is also unappealing considering that a matrix
multiplication can be thought of as a rotation and a scaling. The same rotation at every
element seems unlikely to be useful, and the scaling is exactly what's captured in diagonal
vectors $\lambda_t$. Recurrent coefficient vectors $\lambda_t$ provide enough flexibility
to implement schemes such as exponential moving averages or a gating mechanism.

\subsection{Backpropagation}
\begin{align*}
\nabla_{h_T}L &= \frac{\partial L}{\partial h_T} \\
\nabla_{h_t}L &= \frac{\partial h_{t+1}}{\partial h_t} \odot \nabla_{h_{t+1}} L + \frac{\partial L}{\partial h_t} \\
&= \lambda_{t+1} \odot \nabla_{h_{t+1}} L + \frac{\partial L}{\partial h_t} \\
\nabla_{\lambda_t}L &= \frac{\partial h_t}{\partial\lambda_t} \odot \nabla_{h_t}L = h_{t-1} \odot \nabla_{h_t}L \\
\nabla_{x_t}L &= \nabla_{h_t} L \\
\nabla_{h_0}L &=  \frac{\partial h_1}{\partial h_0} \odot \nabla_{h_1} L = \lambda_1 \odot \nabla_{h_1} L
\end{align*}

The backpropagation equations center around a linear recurrence over $\frac{\partial L}{\partial h_t}$ in the reverse order of the original sequence. This allows for parallelizing both the forwards and backwards pass of a linear RNN over the sequence length.

\subsection{Implementation}
GPUs commonly used for deep learning in 2017 consist of between 640 and 3200 parallel
processors known as warps. Each warp operates on 32 single precision floating
point numbers in parallel.

This work implemented parallel linear recurrence as a CUDA kernel with bindings
into the TensorFlow \citep{abadi2016tensorflow} framework. Each warp acts as a
processor, which means the algorithmic $p$ is up to 3200 and the theoretical
parallelization speedup factor is up to several hundred.  The 32 lanes of each
warp work on different elements of the recurrence vector in parallel. These
implementation details mean that peak performance is only obtained on sequences
of at least several thousand steps on at least a 32 element vector.

The parallel linear recurrence CUDA kernel and TensorFlow bindings are
available at \url{https://github.com/eamartin/parallelizing_linear_rnns} .

\section{Models}
Parallel linear recurrence can be used to construct a wide variety of differentiable modules that can be evaluated in parallel. Common applications of linear recurrence include gating schemes and exponential moving averages. Although linear recurrence values can depend only linearly on previous elements, the stacking of linear recurrent layers separated by non-linearities allows for a non-linear dependence on the past. In this sense the non-linear depth of a linear recurrent network is the number of layers and not the sequence length.

\subsection{Gated impulse linear recurrent layer}
A gated impulse linear recurrent (GILR) layer transforms its $m$ dimensional inputs $x_t$ into a sequence of $n$ dimensional hidden states $h_t$:
\begin{align*}
g_t &= \sigma(Ux_t + b_g) \\
i_t &= \tau(Vx_t + b_z) \\
h_t &= g_t \odot h_{t-1} + (1-g_t)\odot i_t
\end{align*}
A GILR layer applies the same non-linear transform to each sequence element and
then accumulates the sequence elements with a non-linear gating mechanism. Gate
$g_t$ uses the sigmoid activation function to give values in [0,1] for
reasonable gating semantics, while impulse $i_t$ can use any activation function
$\tau$. Stacking GILR layers allows for rich non-linear dependence on previous
events while still taking advantage of fast parallel sequence evaluation.

\subsubsection{Impact on effective "batch size"}
Consider evaluating an RNN with recurrence $h_t = \sigma(Uh_{t-1} + Vx_t + b)$
from $m$ inputs to $n$ hidden units on a sequence of length $T$ with minibatch
size $b$ using a serial evaluation strategy. At each of $T$ iterations, the
naive approach performs two $(b, m) \times (m, n)$ matrix
multiplications. Larger matrix multiplications achieve higher throughput due to
less IO overhead, so the better approach computes $Vx_t$ for all $t$ ahead of
time in a single $(bT, m) \times (m, n)$ matrix multiply. The non-linear
recurrence forces even the better approach to perform $T$ potentially small $(b,
m) \times (m, n)$ matrix multiplications in serial. This makes serial RNN
performance heavily dependent on minibatch size.

Now consider the GILR, noting that it has the same two matrix-vector
multiplications per iteration as the above RNN. The intermediate variables $g$
and $i$ can be evaluated for all $t$ with a single $(bT, m) \times (m, n)$
matrix multiplication each. Given $g$ and $i$, $h$ can be computed using a
parallel linear recurrence over $T$ vectors each of $bn$ elements. Rather than
$T$ small operations, the GILR can be evaluated over all sequence elements with
two large matrix multiplications and a parallel linear recurrence. GILR performance
is much less dependent on batch size as the matrix multiplication kernel sees an "effective
batch size" of $bT$ and $T$ is typically large.

\subsection{Linear surrogate RNNs}
\label{sec:ls-rnns}
RNNs learn a transition function $s_t = f(s_{t-1}, x_t)$ which combines previous
state $s_{t-1}$ with input $x_t$ to compute current state $s_t$. Non-linear $f$
prevents application of the parallel linear recurrence algorithm and forces slow
serial evaluation. To work around this inefficiency, note that $s_t$ serves
dual purposes. In $s_t = f(s_{t-1}, x_t)$, $s_{t-1}$ serves as an input to $f$
summarizing the previous inputs while $s_t$ serves as the output of $f$ to be
passed to other layers of the network. We can decouple these uses and introduce
independent variables for each purpose: \(s_t\) is passed onto other layers of the network
and we introduce the linear surrogate \(\tilde{s}_t\) which is passed onto the
next state, with \(s_t = f(\tilde{s}_{t-1}, x_t)\). We are still able to choose a
non-linear \(f\), our only limitation being that \(\tilde{s}_t\) must be linearly
computable.  We refer to this class of model as a linear surrogate RNN
(LS-RNN). QRNNs \citep{bradbury2017quasi} are LS-RNNs using $\tilde{h}_{t-1}
= W_k x_{t-k} + ... W_1 x_{t-1}$ and strongly typed
RNNs \citep{balduzzi2016strongly} are LS-RNNs with $\tilde{h}_t=x_{t-1}$. Although
not a rule, LS-RNNs can often be parallelized over sequence length with either
convolution or linear recurrence.

Consider an LSTM:
\begin{align*}
  f_t, i_t, o_t &= \sigma(U_{f,i,o} h_{t-1} + V_{f,i,o} x_t + b_{f,i,o}) \\
  z_t &= \tau(U_z h_{t-1} + V_z x_t + b_z) \\
  c_t &= f_t \odot c_{t-1} + i_t \odot z_t \\
  h_t &= o_t \odot c_t
\end{align*}
An LSTM has state $s_t = (h_t, c_t)$. Since $c_t$ depends only
linearly on $c_{t-1}$, no surrogate is needed for $c_t$. $h_t$ has a non-linear
dependence on $h_{t-1}$, so $h_t$ needs a linear surrogate. Introducing a GILR layer as
the surrogate, we obtain the GILR-LSTM:

\begin{align*}
  g_t &= \sigma(V_g x_t + b_g) \\
  j_t &= \tau(V_j x_t + b_j) \\
  \tilde{h}_t &= g_t \odot \tilde{h}_{t-1} + (1-g_t)\odot j_t \\
  f_t, i_t, o_t &= \sigma(U_{f,i,o} \tilde{h}_{t-1} + V_{f,i,o} x_t + b_{f,i,o}) \\
  z_t &= \tau(U_z \tilde{h}_{t-1} + V_z x_t + b_z) \\
  c_t &= f_t \odot c_{t-1} + i_t \odot z_t \\
  h_t &= o_t \odot c_t \\
\end{align*}

For $m$ inputs and hidden size $n$, a GILR-LSTM contains $2n(n+m)$ more
parameters than the equivalently sized LSTM to handle the mapping from $x$ to
$\tilde{h}$. More generally, a LS-RNN contains all of the same parameters as the
underlying RNN as well as some additional parameters to compute the linear
surrogate.

\section{Experiments}
We perform several experiments. First we find that our parallel linear recurrence
kernel is able to achieve up to 40x higher throughput than a serial implementation
when applied to long sequences. Secondly, we confirm that this kernel speedup
translates to up to a 9x speedup to LS-RNNs such as QRNNs.

In order to illustrate that the linearization does not necessarily come at the
cost of expressibility, we show that the GILR-LSTM architecture computed with
the parallel linear recurrence algorithm is able to train significantly faster
than an optimized LSTM implementation on a pathological long-term dependency
problem from the original LSTM paper \citep{hochreiter1997long}.

\subsection{Throughput benchmarks}

\subsubsection{Kernel performance}
We first illustrate the throughput advantage of the parallel scan algorithm for
evaluating the linear recurrence. For a minibatch comprised of \(b\) sequences
of length \(T\), we define the number of events as \(bT\) and the throughput as
the number of events processed per second. We implement two CUDA kernels, one which
evaluates the parallel linear recurrence described in algorithm \ref{alg:plr},
and one which evaluates the same linear recurrence on GPU in serial over sequence length
and in parallel over features and minibatch. The performance of each kernel depends on
two factors: the sequence length and the product of number of features and minibatch size.
The performance measurements for this experiment are made directly at the kernel level,
avoiding any overhead from TensorFlow. We find that the parallel kernel has
a distinct advantage at long sequence lengths with a speedup factor of up to 40x,
as shown in table \ref{table:kernel-throughput}.
The parallel kernel does not perform well at short sequence lengths due to the
overhead of multiple passes over data and communication between processors.
\begin{table}[t]
\caption{Parallel kernel speedup on $m$ features (minibatch size $= 1$)}
\begin{center}
\begin{tabular}{@{}llll@{}}
\label{table:kernel-throughput}
Sequence Length & \(m=4\)  & \(m=32\) & \(m=128\) \\ \midrule
16              & 0.06 & 0.06 & 0.05  \\
256             & 0.22 & 0.22 & 0.86  \\
4,096           & 1.02 & 2.94 & 3.36  \\
65,536          & 38.5 & 41.8 & 17.5  \\ \bottomrule
\end{tabular}
\end{center}
\end{table}

\subsubsection{Accelerating existing RNN architectures}
Several recently introduced LS-RNNs can be accelerated with the parallel linear
recurrence algorithm. We implemented SRUs, QRNNs (with filter width 2 and 10),
and GILR-LSTMs that can be computed with either the standard serial linear
recurrence algorithm or parallel linear recurrence. Both methods compute an
identical recurrence, so switching from a serial to parallel implementation does
not cause any numerical changes and takes only a single line of code changes.
Notably, both SRUs and QRNNs claim an order of magnitude speedup compared to
CuDNN LSTM when implemented with serial linear recurrence. Any further speedup
from parallel linear recurrence applies on top of the existing speedup.  We
timed train throughput (forwards and backwards propagation), but the linear time
of each pass also makes the results applicable to forwards (inference)
performance. However, parallel linear recurrence can only accelerate inference
in scenarios where the entire input sequence is known at the start of the
inference phase.  We controlled for GPU memory usage within these experiments by
fixing $bT = 65,536$ for minibatch size $b$ and sequence length $T$, and chose a
popular architecture consisting of two stacked RNN layers with 256 hidden units
and an input size of 4.

Table \ref{table:rnn-throughput} shows that the
throughput advantage from using parallel linear recurrence compared to
serial linear recurrence reaches up to 9x. Simpler architectures
(for which the linear recurrence is a higher proportion of the
total computational load) are more affected by the switch to the parallel kernel.
This is
particularly clear in the case of the QRNN, where including wider
convolutional filters results in more time spent outside of the linear recurrence and
therefore reduces the speedup from linear recurrence parallelization.
\begin{table}[t]
  \caption{Parallel kernel speedup for a variety of LS-RNNs, implemented as two
    stacked RNN layers with 256 hidden units. We keep the GPU memory usage constant
  by fixing $bT = 65,536$ for minibatch size $b$ and sequence length $T$}
\begin{center}
\begin{tabular}{@{}lrrrr@{}}
\label{table:rnn-throughput}
Sequence Length & SRU & QRNN (filter size 2) & QRNN (filter size 10) & GILR-LSTM\\ \midrule
16 & 0.28 & 0.38 & 0.78 & 0.61\\
256 & 0.84 & 0.86 & 0.99 & 0.91\\
4,096 & 1.38 & 1.18 & 1.05 & 0.98\\
65,536 & 9.21 & 6.68 & 2.05 & 1.41\\ \bottomrule
\end{tabular}
\end{center}
\end{table}








\subsection{Synthetic Experiment}
One of the key strengths of the LSTM is that it is capable of dealing with
long-term dependencies. In order to demonstrate that the GILR-LSTM is also able
to handle long-term dependencies we tackle a canonical example of inference over
many time steps from \citet{hochreiter1997long}.  We show that in fact the
GILR-LSTM is able to outperform the CuDNN LSTM and extend to sequence lengths
orders of magnitude longer than dealt with previously. The input consists of
sequences of length $n$ where for $n > 0$ each element is a randomly chosen
one-hot vector \(x\) in $p$-dimensional space.  The first vector in each
sequence, \(x_0\), is always either \((1, 0, \ldots, 0)\) or \((-1, 0, \ldots,
0)\). The sequential model must read in an entire sequence and then output the
sign of the first sequence element.  This sequence classification problem
requires remembering the first element over the length of the sequence, and
early RNNs struggled with this for \(p\) as small as a few dozen. In the
original formulation of the problem (dealing in the regime with around one
hundred timesteps), the dimensionality of the input \(p\) is set equal to
\(n\). Since this would make the size of the input data grow impractically large
as \(\mathcal{O}(n^2)\) for long sequences, we fix \(p = 128\) as we vary \(n\).
\begin{figure}[] \centering
\includegraphics[width=12cm]{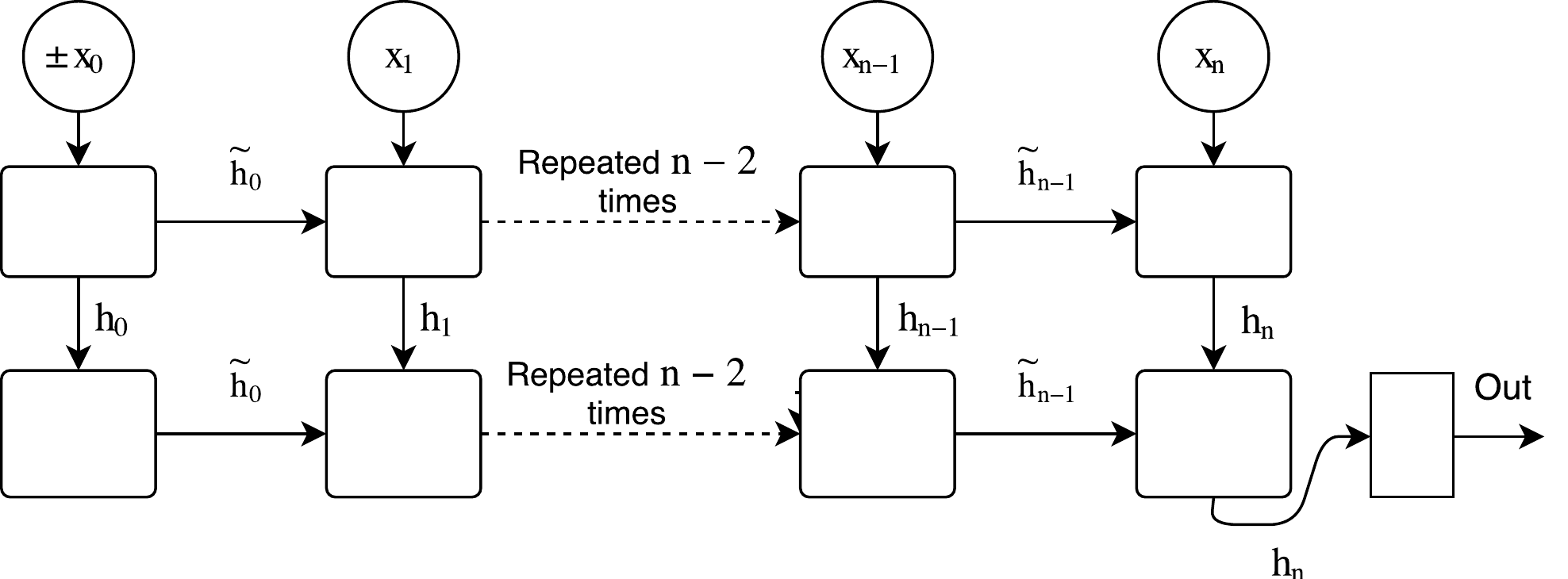}
\caption{The structure of the synthetic example and the GILR-LSTM architecture
we used to tackle it. We feed in one-hot unit vectors \(x\) which are chosen
uniformly at random (with replacement). The class is determined by the very
first vector \(x_0\), which has a fixed direction. The sign of \(x_0\)
determines the class. In the diagram, each rounded block indicates a cell of the
RNN, whilst the square indicates a linear unit.}
\label{fig:synthetic_diagram}
\end{figure} We generated sequences for $n$ equal to 1,024, 8,192, and
1,048,576. For each of these we compared a two layer GILR-LSTM with 512 hidden
units to a two layer LSTM with 512 hidden units\footnote{For the longest
sequence length, the number of hidden units was decreased to 64 for both
architectures so that the net could fit in memory.} per layer implemented by
CuDNN.

\begin{figure}
  \centering
  \begin{minipage}{0.5\textwidth}
    \includegraphics[width=1.0\textwidth]{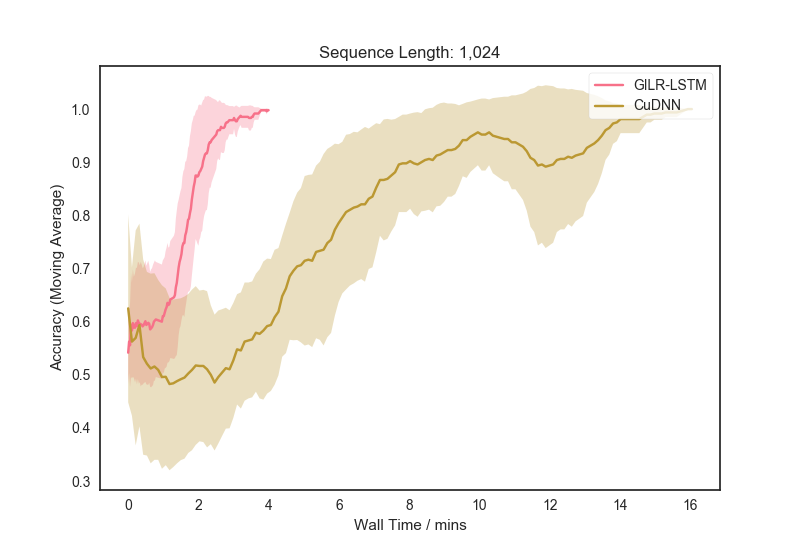}
  \end{minipage}%
    \begin{minipage}{0.5\textwidth}
    \includegraphics[width=1.0\textwidth]{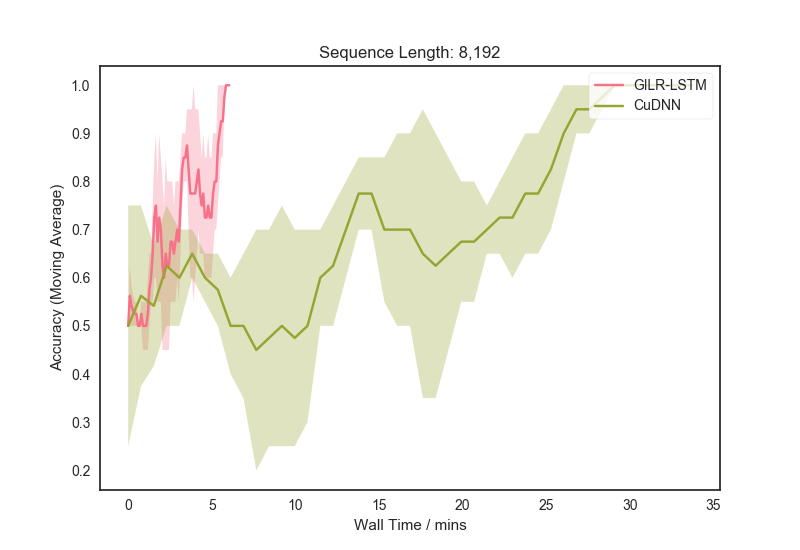}
  \end{minipage}
  \begin{minipage}{0.5\textwidth}
    \includegraphics[width=1.0\textwidth]{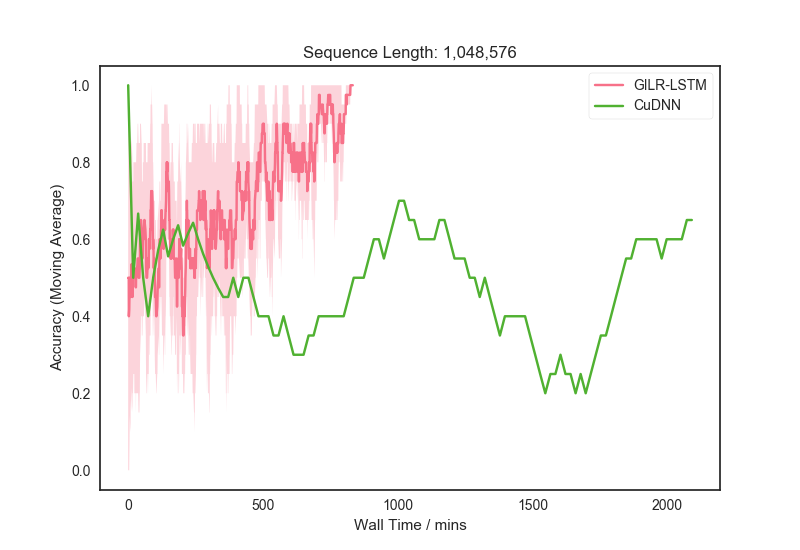}
  \end{minipage}
  \caption{Learning curves for GILR-LSTM and CuDNN LSTM architectures for various
    sequence lengths. Each plot shows the moving mean and standard deviation of
    classification accuracy over five training runs, with the exception of a single
    run for CuDNN LSTM on 1 million sequence length.}
    \label{fig:synthetic_training}
  \end{figure}

We ran all experiments on a NVIDIA K80 GPU, with five runs per configuration
allowing us to find the average and standard deviation of the time and number of
iterations to convergence. We continually generated random sequences to serve
as input data.  A brief search over learning rate
and batch size was carried out to find the parameters which allow the network to
converge most rapidly for all runs. The criterion for convergence was five consecutive
minibatches giving 100\% accuracy. The learning curves
in figure \ref{fig:synthetic_training} give support to this being a reasonable
convergence criteria.
For the longest sequence length, we did not observe the CuDNN LSTM converging, even after
several days' training.

\begin{table}[]
\label{table:synth-table}
\centering
\caption{Performance of the GILR-LSTM compared to the CuDNN LSTM
  on problem 2b from \citet{hochreiter1997long}. }
\begin{tabular}{@{}lllllll@{}} \toprule Sequence Length &
\multicolumn{2}{c}{\textbf{1,024}} & \multicolumn{2}{c}{\textbf{8,192}} &
\multicolumn{2}{c}{\textbf{1,048,576}} \\ \midrule &CuDNN& GILR &
CuDNN &GILR&CuDNN&GILR\\ \cmidrule(l){2-7} Iterations (1000s) &
                     1.0 
                     \(\pm\) 0.4 & 0.55 
                                    \(\pm\) 0.04 & 0.44 
                                                   \(\pm\) 0.05 & 0.56 
                                                                 \(\pm\) 0.16
                                                               &-& 14 \(\pm\) 3 \\
\begin{tabular}[c]{@{}l@{}}Wall time (hours)\end{tabular} &
0.28 \(\pm\) 0.08 & 0.031 \(\pm\) 0.002 & 0.58 \(\pm\) 0.06 & 0.10 \(\pm\) 0.03
&-& 9.7 \(\pm\) 1.7 \\ \bottomrule
\end{tabular}
\end{table}

The results as show in table \ref{table:synth-table} illustrate that the GILR-LSTM is able to converge between 6 and 10
times faster than the CuDNN LSTM. This is somewhat surprising
given the LSTM was specifically constructed for problems of this sort, and the
CuDNN LSTM implementation is highly optimized (to the extent that
the monolithic interface it exposes is difficult to modify or extend). The
GILR-LSTM is implemented entirely in standard TensorFlow with the exception of
using the new linear recurrence op instead of a TensorFlow symbolic loop.
Convergence of the GILR-LSTM models
leads to the conclusion that the non-linearities present in LSTM are
not necessary for solving this instance of the long-term dependency problem.
The time to convergence further leads to the conclusion that inclusion of a
non-linearity at every step incurs a significant training time slowdown.
Furthermore, the GILR-LSTM is able to learn to carry dependencies over a one
million element sequence. As far as we know, this one million step sequence
experiment is the longest sequential learning problem to be handled by neural
networks to date.

\section{Conclusion}
A significant portion of the success of deep learning can be attributed to access to massive
amounts of computation. Most of this computation is accessed through two
highly efficient and parallelizable building blocks:
matrix multiplication and convolution. Recent research has demonstrated that
linear RNNs can achieve similar prediction accuracy to non-linear RNNs on a wide
variety of tasks in a fraction of the training time. We propose the framework of
LS-RNNs as a way to tame the growing zoo of sequential neural nets. We identify linear
recurrence as another parallelizable building block for current and future sequential models
and we use it to obtain significant speedups on already fast models.
With the power of parallel
linear recurrence we are able to solve a sequential dependency problem multiple
orders of magnitude larger than anything done prior.
Future applications of parallel linear recurrence within neural nets could include
parallel training of memory augmented models or providing a new sort of image filter
on very high resolution images.
We hope that parallel linear recurrence can be to large scale sequence
modelling what fast convolution algorithms are to image recognition.

\subsubsection*{Acknowledgments}
We would like to acknowledge Kevin Bowers, Alex Meiburg, JD Co-Reyes, Carson
McNeil, Andy Palan, Sören Mindermann, and several others for fruitful conversations and guidance.

\end{document}